# LLM-Assisted Formalization Enables Deterministic Detection of Statutory Inconsistency in the Internal Revenue Code

Borchuluun Yadamsuren,[1,*] Steven Keith Platt,[2] Miguel Diaz[3]

[1, 2, 3] Laboratory for Applied Artificial Intelligence, Quinlan School of Business, Loyola University Chicago, Chicago, IL, USA
byadamsuren@luc.edu; splatt1@luc.edu; mdiaz37@luc.edu
[*]Correspondence: byadamsuren@luc.edu

**Abstract**
This study introduces a hybrid neuro-symbolic framework that achieves deterministic detection of statutory inconsistency in complex law. We use the U.S. Internal Revenue Code (IRC) as a case study because its complexity makes it a fertile domain for identifying conflicts. Our research offers a solution for detecting inconsistent provisions by combining Large Language Models (LLMs) with symbolic logic.

LLM-based methods can support compliance, fairness, and statutory drafting, yet tax-specific applications remain sparse. A key challenge is that such models struggle with hierarchical processing and deep structured reasoning, especially over long text.

This research addresses these gaps through experiments using GPT-4o, GPT-5, and Prolog. GPT-4o was first used to translate Section 121 into Prolog rules and refine them in SWISH. These rules were then incorporated into prompts to test whether Prolog-augmented prompting improved GPT-4o's inconsistency detection. GPT-4o, whether prompted with natural language alone or with Prolog augmentation, detected the inconsistency in only one of three strategies (33 percent accuracy), but its reasoning quality differed: natural-language prompting achieved 100 percent rule coverage, while Prolog-augmented prompting achieved 66 percent, indicating more incomplete statutory analysis.

In contrast to probabilistic prompting, the hybrid Prolog model produced deterministic and reproducible results. Guided by GPT-5 for refinement, the model formalized the IRC section's competing interpretations and successfully detected an inconsistency zone. Validation tests confirm that the Prolog implementation is accurate, internally consistent, deterministic, and capable of autonomously identifying inconsistencies. These findings show that LLM-assisted formalization, anchored in symbolic logic, enables transparent and reliable statutory inconsistency detection.

**Introduction**

AI is being deployed in many fields, including in the legal domain. For example, Bernsohn et al. (2024) leveraged LLMs to identify legal violations in unstructured text, demonstrating the potential for LLMs to handle complex legal classification tasks. Similarly, Bui et al. (2024) utilized LLM-based data augmentation techniques to address data scarcity in legal datasets, significantly improving entailment and retrieval tasks. Despite these advances, LLMs often struggle with precise and consistent legal reasoning due to their probabilistic nature and inability to operate on explicit logical structures. This has led to increasing interest in neuro-symbolic systems that combine LLMs with deterministic reasoning tools. Numerous studies have explored using Prolog, a widely adopted logic programming language (Körner et al., 2022), to complement LLMs in addressing these reasoning limitations.[1] For example, Sehgal and Liu (2025) developed the LogicLease system for analyzing landlord-tenant legal cases by incorporating a Prolog component. Zin et al. (2025) investigated prompting strategies to formalize traffic rules in Prolog using LLMs.

Notwithstanding, the use of LLMs to assist in tax-related investigations is particularly challenging. Oei & Osofsky (2018) note, for example, that U.S. tax laws are complex, often ambiguous, and not particularly intuitive. This body of law is also characterized by highly technical language and intricate cross-references. In addition, it is also complicated and ever-evolving. For instance, Oei & Osofsky (2018) conducted in-depth interviews with government attorneys who participated in the tax legislative process over four decades. They identified several compounding issues, such as the accretion of statutory language over time, limitations on changing existing statutory language, time pressure while drafting tax legislation, low likelihood of technical corrections, and a lack of prioritization to clean up the tax code.

Lawsky (in press) has also recognized the complexity of tax law drafting by pointing out an inconsistency and contradiction in the tax code. Extending this work, we examine whether AI-assisted formalization can support systematic inconsistency detection. This study addresses two related questions. First, can GPT-4o, when prompted with natural language and Prolog-augmented prompts, reliably detect a known inconsistency in IRC §121? Second, can a hybrid approach, in which an LLM assists in generating and refining formal Prolog representations of statutory text, produce a deterministic inconsistency detection system that reliably identifies conflicts across a diverse set of use cases? To explore these questions, we devise and test a system that utilizes GPT-4o and GPT-5 for prompt augmentation and rule formalization, while relying on deterministic symbolic reasoning for inconsistency detection executed in Prolog.

---

[1] Prolog (Programming in Logic) is a declarative programming language based on first-order predicate logic, widely used for symbolic reasoning and knowledge representation. In Prolog, computation occurs through pattern matching and logical inference over facts and rules, making it well-suited for tasks involving natural language understanding, expert systems, and automated reasoning.

The study's focus is on ambiguous language contained in IRC Section: *'whichever is applicable,'* which can produce divergent outcomes under the same fact pattern (Lawsky, in press). We formalize both interpretations that create this problem and test them computationally. This computational approach is, to our knowledge, the first applied to uncover statutory conflicts in the tax code utilizing an LLM-assisted formalization.

This computational method combines a large language model for translating tax law into formal logic with a Prolog-based symbolic reasoner to detect inconsistency cases. Prolog offers a widely used, standards-backed platform that lowers adoption friction and supports reproducible legal reasoning (Körner et al., 2022).

In addition to detecting tax inconsistencies in existing law, an additional void exists in developing an automated and scalable solution for detecting inconsistencies during the legislative drafting process itself. This issue is not merely academic, as it has practical implications. As noted above, undetected statutory inconsistencies can lead to compliance issues, potentially unfair outcomes, and erosion of public trust in the tax system. Such contradictions represent a failure of the legal system to provide clear guidance. The development of an automated detection solution to aid in legal drafting is therefore useful for creating a more efficient, fair, and predictable legal framework.

The contributions of this study include:

- Combining LLMs with symbolic logic through a systematic evaluation for statutory inconsistency detection.
- Develop a replicable experimental framework enabling transparent, auditable legal reasoning with full reproducibility.
- Propose a novel methodology to measure the reasoning performance of LLMs in uncovering inconsistencies in tax statutes.
- Present a system that can provide feedback to legislative counsel, flagging potential inconsistencies during the legislative drafting process.

This research is organized as follows. Section I reviews the relevant literature on legal inconsistencies and AI-driven statutory analysis. Section II details our research methodology, including our hybrid framework and evaluation rubric. Section III presents the empirical findings from our comparative experiments. Finally, Section IV discusses the implications of our results, acknowledges the study's limitations, and proposes directions for future research.

## I. Literature Review

The Internal Revenue Code is notoriously complex (Choi, 2020). Taxpayers face challenges both in interpreting and combining disparate provisions of the IRC. Choi (2020) highlights how judicial application of anti-abuse doctrines, for example, creates additional inconsistency and confusion, particularly when substance-over-form doctrines conflict with provisions that are purely formal in nature.

Oei & Osofsky (2018) document how Congress has extensively modified tax legislation over the years, with the speed of passing new legislation drawing protests from experts due to the resulting complexity. They advocate for a complete paradigm shift in the legal drafting process and specifically highlight the importance of using modern technology, including artificial intelligence, to optimize tax law drafting and analysis. Lawsky (in press) suggests that formalizing statutes in programming language can serve as a safeguard against writing inconsistencies into the Code, while simultaneously improving transparency and accuracy.

Framework for Legal Inconsistency

Donelson (2019) provides a comprehensive framework for understanding inconsistencies and ambiguities in the law, isolating four types. These include irreconcilable inconsistencies between laws, reconcilable inconsistencies between laws, inconsistencies between justifications of laws, and inconsistencies between justification and execution. Building on this framework, Lawsky (in press) applied Donelson's definition of irreconcilable inconsistency and identified a case that yields inconsistent outcomes under the same set of facts in IRC §121[2] by isolating the interaction between §121(b)(2)(A), §121(b)(2)(B), and §121(c). This inconsistency arises when married couples file joint tax returns and partially qualify for a home sale exclusion. The statute instructs taxpayers to apply reductions to "the dollar limitation under paragraph (1) or (2) of subsection (b), *whichever is applicable*," but fails to clarify which limitation applies when couples have asymmetric qualifications.

This creates scenarios where the same facts can yield different legal outcomes, depending on interpretation, which is a classic example of irreconcilable inconsistency.[3]

Large Language Models in Legal Reasoning

Recent research has explored LLM capabilities in legal contexts with mixed results. Nay et al. (2024) tested various LLMs on multiple-choice tax law questions. They found that while giving models more legal text slightly increased accuracy, Chain of Thought (CoT) prompting did not consistently improve performance across models. They concluded that the models fell short of expert-level performance in tax law applications.

---

[2] Section 121 of the IRC addresses the exclusion of gain from the sale of a principal residence.
[3] It is noteworthy that this inconsistency exists in the statute as written, but has since been resolved by IRS regulations, consistently applying the sum of limitations approach. 26 CFR § 1.121-2(a)(3) (ii) reads, in part: "For taxpayers filing jointly, if either spouse fails to meet the requirements of paragraph (a)(3)(i) of this section, the maximum limitation amount to be claimed by the couple is the sum of each spouse's limitation amount determined on a separate basis as if they had not been married."

Fei et al. (2023) developed the LawBench evaluation framework, assessing LLMs across three cognitive levels: legal knowledge memorization, understanding, and application. Their extensive evaluation of fifty-two LLMs found GPT-4o to be the best-performing model in legal domains, though significant limitations remained in complex reasoning tasks due to the inherent limitations of generative AI models in handling the intricate cognitive tasks required for legal reasoning. Similarly, Blair-Stanek et al. (2023) observed frequent failures in GPT-3's statutory reasoning due to incomplete or inaccurate prior knowledge of statutory content and inherent reasoning limitations. Guha et al. (2023) developed a legal reasoning benchmark based on the empirical evaluation of 20 open-source and commercial LLMs by testing 162 different tasks covering six types of legal reasoning. They found significant variation in the performance of LLMs across tasks.

Kant et al. (2025) emphasized that recent LLMs such as OpenAI o1 have demonstrated notable improvements in legal reasoning, with newer models like DeepSeek R1 and OpenAI o3-mini, achieving comparable results at lower computational cost. These developments suggest that reasoning-oriented LLMs hold the promise for automating complex legal tasks.

Symbolic Logic and Hybrid Approaches in Legal AI

Lawsky (2017) advanced the idea that translating proposed statutory language into logical terms, or formalizing legislation, can prevent unintentional ambiguity and lead to a computationally legible legal code. Building on this foundation, Lawsky (2022) argues that tax law is already "formalized" in practice through forms, worksheets, and software, but often in ways that obscure the underlying legal structure. For a more transparent alternative, she introduced Catala, a domain-specific programming language that encodes tax statutes as prioritized default logic in which general rules are overridden by exceptions. This solution mirrors how tax law is drafted and requires explicit priority choice when rules might clash.

Further, Lawsky (in press) developed an inconsistency-checking pipeline for I.R.C. §121 using Python and Catala. She implements two formalizations: a per-taxpayer "sum of limitations" reading that mirrors IRS Publication 523,[4] and a joint-cap reading that scales the $500,000 limit by a joint numerator. Using Python with the Z3 theorem prover (de Moura & Bjørner, 2008), she encodes the statute's temporal tests and dollar limits, then systematically explores ownership, use, prior exclusions, and gain to identify fact patterns where the two readings produce different results. In parallel, Catala is used prescriptively, as it relies on prioritized default logic. Potentially conflicting rules must be ordered in advance, which forces an explicit resolution, whereas Z3 highlights scenarios where the statute leaves key conditions underspecified or ambiguous. Catala attempts to avoid such ambiguity by requiring the programmer to make the legal priority explicit before execution. Together, these methods move formalization from concept to practice by

---

[4] IRS Publication 523, Selling Your Home, provides the tax rules and guidance for taxpayers who sell or exchange their primary residence. The focus of this publication is to explain the rules surrounding the exclusion of capital gain from the sale of a primary residence.

enabling operational inconsistency checking, with open code that makes the modeling choices transparent and reproducible (Lawsky, in press; Lawsky, n.d).

Holzenberger et al. (2020) established foundational research in computational statutory reasoning by developing the Statutory Reasoning Assessment (SARA) dataset, focusing explicitly on U.S. law. Their work highlighted a clear distinction between probabilistic approaches using LLMs such as BERT and Legal BERT, and deterministic symbolic methods using the Prolog language. While probabilistic models showed flexibility, they performed poorly on precise legal reasoning tasks, frequently failing to capture the nuanced logical structures inherent in statutory language. Conversely, the deterministic Prolog-based solver demonstrated complete accuracy and interpretability, effectively addressing the logical deficiencies observed using LLMs. They concluded that integrating LLMs with symbolic or structured logical systems, such as Prolog or other structured reasoning tools, could improve logical rigor and consistency in legal reasoning tasks.

Holzenberger and Durme (2023) refined this approach by proposing hybrid methodologies combining automated information extraction (IE) from natural language with deterministic Prolog-based symbolic reasoning. This approach aimed to reduce the substantial human effort previously needed for the manual translation of cases into Prolog logic. By evaluating various pre-trained NLP models, they demonstrate that automated IE methods, especially those utilizing domain-adapted models such as LEGALBERT, could significantly improve downstream symbolic reasoning accuracy. However, automated IE still encountered challenges, such as missing critical elements, including monetary values or specific dates, leading to diminished reasoning accuracy.

Kant et al. (2024) emphasized the importance of differentiating deterministic and non-deterministic components of legal reasoning before applying AI. They successfully translated laws and insurance contracts into Prolog rules using LLMs, finding that OpenAI's o1-preview model significantly outperformed GPT-4o in generating accurate Prolog queries. In addition, Kant et al. (2025) evaluated the effectiveness of combining LLMs with Prolog-based logic encoding by comparing three approaches: direct (vanilla) LLM prompting, unguided Prolog generation, and expert-guided Prolog generation based on a pre-defined and validated set of insurance claim coverage questions and answers from two U.S. health insurance policies. Under the expert-guided approach, the GPT-4o and o1 models performed with 100% accuracy for the simplified version of the insurance contract. For the more complex insurance policy, the OpenAI o1 model demonstrated higher accuracy than other models. The Prolog code was evaluated using the SWISH interpreter, highlighting the value of symbolic logic for generating interpretable and auditable legal outputs.

**Research Gap and Contributions**

Despite these advances, limited research exists on using LLMs specifically for detecting inconsistencies within a complex statutory framework like tax law. This research makes four distinct contributions to the AI legal literature:

1. Empirical Performance Comparison: The first systematic comparison of an LLM comparing its performance with an LLM-enhanced symbolic logic approach for statutory inconsistency detection is provided, revealing a performance gap based on output accuracy.

2. Hybrid Workflow Development: A practical pipeline is established that optimizes each technology by positioning LLMs as formalization assistants rather than as primary reasoning engines, leveraging each technology's strengths.

3. Methodological Framework: A replicable experimental framework using the SWISH environment is developed. This enables transparent, auditable legal reasoning with complete reproducibility.

4. Performance Evaluation: A novel methodology is introduced to measure LLM reasoning performance in uncovering inconsistencies in tax statutes.

In addition, we highlight the important application of this methodology in integrating into the legislative drafting process itself. That is, we lay the groundwork for a system that can provide feedback to legislative counsel, flagging potential inconsistencies during the draft writing process, which could prevent these issues from becoming law.

**Research Questions**

This investigation centers on several distinct yet complementary research questions.

Research Question One (RQ1): Can GPT-4o, when prompted with natural language and Prolog-augmented prompts, reliably detect a known inconsistency in IRC §121?

Research Question Two (RQ2): Can a hybrid approach, in which an LLM assists in generating and refining formal Prolog representations of statutory text, produce a deterministic inconsistency detection system that reliably identifies conflicts across a diverse set of use cases?

## II. Research Methodology

We develop a methodology for detecting a known inconsistency in IRC §121 by combining large language models (LLMs) with deterministic symbolic logic. To accomplish this, a two-phase experimental design is formalized. Phase One investigates whether GPT-4o, when prompted directly in natural language and with Prolog-augmented prompts, can reliably detect the known inconsistency, thereby testing the limits of probabilistic reasoning alone. Phase Two evaluates whether LLMs can generate and refine Prolog code that generalizes across a diverse set of use cases. Further, we investigate whether this independently executed implementation in SWISH can detect statutory inconsistencies, thereby assessing Prolog's viability as a formalization framework. This design enables a structured comparison between probabilistic LLM reasoning and deterministic symbolic approaches.

The specific section of the IRC analyzed is §121. In summary, IRC §121 provides that:

Under §121(b)(2)(A), referred to here as the *joint cap*, married couples filing jointly may exclude up to $500,000 of gain from the sale of a principal residence if specific ownership and use requirements are met. If those requirements are not satisfied, §121(b)(2)(B), referred to here as the *sum of limitations*, directs the taxpayer to use the combined total of each spouse's individual $250,000 exclusion. When couples qualify for a reduced exclusion under §121(c) due to unforeseen circumstances, such as relocation for a job, the statute instructs the taxpayer to apply the reduction to "the dollar limitation under paragraph (1) or (2) of subsection (b), whichever is applicable" (§121(c)(1)). However, the statute does not clarify which limitation, the $500,000 joint cap or the sum of limitations reduced amount, is "applicable" when a couple partially qualifies. This ambiguity introduces the *applicable limitation* problem that determines which cap governs in mixed qualification cases and produces what Lawsky (in press) describes as a *condition failure*, where §121(b) and §121(c) yield conflicting results depending on how the limitation is applied.

For purposes of this analysis, a known inconsistency in IRC §121 serves as the ground truth test case (Lawsky, in press). This is presented as follows:

Scenario: H (husband) has owned a house used as his principal residence since 1996. He marries W (wife) on January 15, 1999, and W begins using the house as her principal residence. On January 15, 2000, H sells the house due to a change in W's place of employment.

Legal inconsistency: The statute's phrase "*whichever is applicable*" in §121(c)(1) creates ambiguity about whether the couple should receive a $500,000 joint exclusion or individual exclusions totaling $375,000, leading to irreconcilable interpretations.

Experimental Design

RQ1 Discovery

We first consider Chain-of-Thought (CoT) prompting with and without Prolog reasoning. We adhere to both Wei et al. (2022) and Blair-Stanek et al. (2023) guidance on applying a structured methodological approach that guides LLMs explicitly through intermediate reasoning steps that mirror human legal analysis. This technique is particularly relevant for statutory interpretation tasks, where the reasoning process consists of identifying applicable provisions, analyzing their interaction, and drawing logical conclusions, which is as important as the final determination. *Figure 1* provides a graphical overview of the experimental design for RQ1.

*Figure 1. Experimental Design for RQ1.*

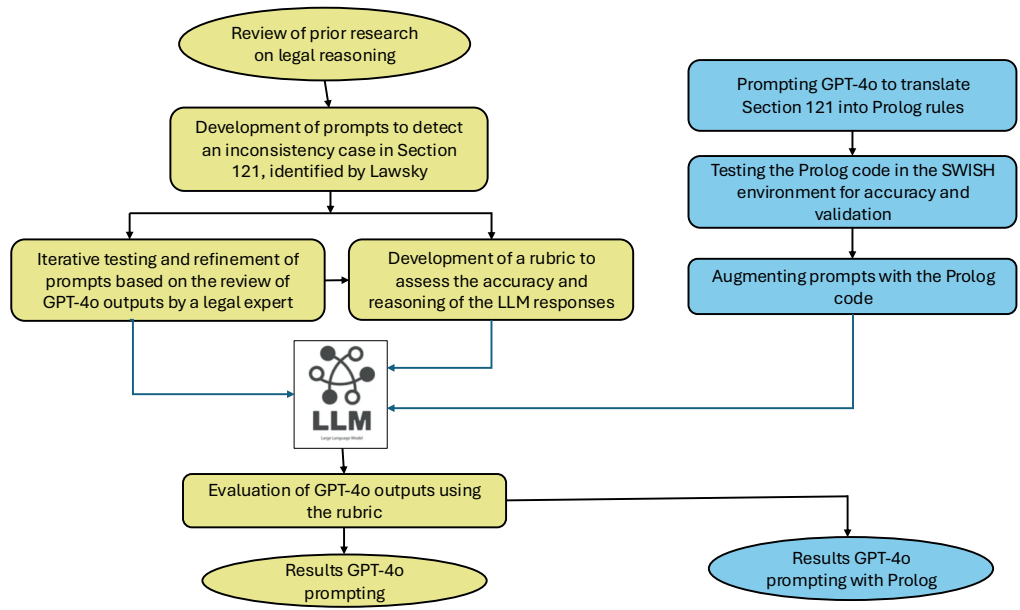

The objective here is to test GPT-4o's ability to detect an inconsistency within Section 121 of the IRC. To achieve this, a series of CoT prompts were developed to guide the logical reasoning process. These prompts were rigorously tested to determine how the guidance from more general to specific prompting could improve the accuracy of the responses from the LLM. Prompts were continuously refined based on accuracy checks by a legal expert.

Before finalizing the experimental protocol, pilot testing was undertaken, in which the selected prompts were run multiple times through the ChatGPT web interface. The output demonstrated a high degree of consistency in underlying reasoning across instances. The only differences observed were stylistic, primarily in word choice and sentence structure, rather than substantive changes in logical steps or conclusions. These stylistic variations are characteristic of generative AI models, but did not affect the reasoning quality relevant to this study. Based on this observed stability, it was determined that running each prompt once per test scenario was sufficient for the main experiment. This design choice balances methodological rigor with practical relevance, reflecting how practitioners might typically interact with the ChatGPT web interface in practice.

An evaluation rubric was developed that enabled a systematic assessment of the accuracy and legal reasoning (rule coverage score) of the responses generated by the LLM, both with and without Prolog augmentation. Three prompts were selected (*Table 1*) for both experiments. The only difference between the two prompting methods was the Prolog augmentation.

All testing was conducted through the ChatGPT web interface, which does not allow setting or controlling system-level parameters such as temperature and token limits. This choice was deliberate. By using the web interface, we replicate how legal practitioners would most likely interact with GPT-4o in practice, without requiring technical expertise, code, or API keys. This enhances replicability for non-technical audiences, while at the same time limiting experimental control over randomness. In addition, the prompt engineering process involved iterative refinement guided by a legal expert. While this human-in-the-loop method improved prompt specificity, we recognize that it may introduce expert bias.

<u>Prompting Without Prolog</u> (Figure 1. Yellow Path- LLM Only): GPT-4o was prompted using the three prompts (*Table 1*, column "Prompt without Prolog"). These prompts did not include any formalized logical structure, allowing an evaluation of the model's baseline reasoning performance to detect legal inconsistencies.

<u>Prompting with Prolog</u> (Figure 1. Blue Path- LLM + Prolog Augmentation): this encompassed the following procedure:

1. GPT-4o was prompted to translate Section 121 into Prolog rules. The prompts did not provide structural guidance or clause breakdown. This step was designed to test the LLMs ability to formalize legal text without human guidance. The model's response formed a preliminary Prolog rule set. GPT-4o was further prompted to check the accuracy of the generated Prolog rules (the model indicated that its output did not capture all rules presented in §121). To refine and validate the generated Prolog rules by GPT-4o, the Prolog SWISH environment was used for iterative debugging and testing. SWISH executed the initial Prolog rules, flagged syntax and type/instantiation errors, and provided tracing tools to control flow and failure points. The SWISH diagnostics and failing clauses were then fed back into GPT-4o, which proposed targeted fixes. This iterative loop produced a minimal working version of Prolog code (*Appendix 1*) to augment the prompting.
All Prolog code versions referenced in *Appendices 1–3* is available at: https://github.com/borchuluun/section121-inconsistency-detection.

2. The finalized Prolog rules from the above step augmented a set of CoT prompts. The three prompts (*Table 1*, column Prompt with Prolog) explicitly reference the Prolog code and the model to search for inconsistencies in §121 of the IRC to check if the logic derived from the Prolog-based rules could improve the reasoning of the LLM.

RQ1 Evaluation Rubric

Building on the LEGALBENCH framework (Guha et al., 2023), a two-dimensional scoring system of LLM-generated outputs was designed. First, we present the accuracy score, which quantifies overall performance. Second, representing a central methodological contribution of this study, is the development of the rule coverage score. This is significant because in legal analysis, a correct definitive answer (e.g., identifying an inconsistency) may be insufficient if the underlying legal justification is incomplete or flawed (Guha et al., 2023). The rule coverage score thus measures this competency by evaluating whether the AI model identifies and applies all relevant statutory provisions necessary to reason over the inconsistency. This moves beyond simple binary accuracy and provides a granular, qualitative assessment of the model's reasoning process, mirroring the standards of human legal analysis. It distinguishes between what may be a correct guess and a well-reasoned conclusion, ensuring that the evaluation captures not just what the model concludes, but how it reaches that conclusion.

1. Accuracy Score (0/1)

This is a measure of the correct classifications of tax Code interpretation and application, encompassing:

- Issue-spotting: Did the system identify the correct tax problem?
- Rule application: Were the tax Code sections logically applied to the facts?
- Rule conclusion: Was the outcome valid and well-supported?
- Interpretation quality: Were ambiguous terms handled appropriately?

2. Rule Coverage Score (%)

This metric calculates the ratio of correctly identified and applied tax Code sections. The rule coverage score is defined as:

**Rule Coverage Score (%) = (Number of relevant rules correctly identified *and* applied by the LLM ÷ Total number of expected rules) × 100.**

"Identified" means that the statute (e.g., §121(a)) is explicitly cited or clearly referenced.

"Applied" means that the rule is used appropriately and meaningfully in the reasoning process, not just mentioned. Only legal rules or sections contained in IRC §121 that meet both conditions are included in the numerator. This directly aligns with rule-recall (did the model cite the correct sections?) and rule-application (did it apply to each rule mentioned?) The total number of expected rules (denominator) is determined in advance by a subject-matter expert based on the factual scenario.

All responses were scored by a human evaluator with legal training to ensure an accurate assessment of both statutory interpretation and factual application. The ground truth case requires coverage of three key sections: §121(b)(2)(A), §121(b)(2)(B), and §121(c).

RQ2 Discovery

To address Research Question 2, a two-phase methodology was developed to evaluate the viability of LLM-assisted formalization of IRC §121 for statutory inconsistency detection.

In the first phase, using GPT-4o, the Prolog code developed in addressing Research Question 1 (Prompting with Prolog; *Appendix 1*) was further refined in the SWISH environment. The objective of this phase was to ensure that the Prolog code could reproduce and detect inconsistencies across a variety of fact patterns generated with the assistance of GPT-4o. These additional scenarios encompassed different combinations of ownership duration, use periods, and prior-exclusion intervals under §121 and were added as facts to the Prolog code (*Appendix 2*).

In the second phase, the enhanced Prolog code generated from Phase 1 (*Appendix 2*) was further enhanced, this time with GPT-5. GPT-5 was utilized as a computational collaborator, identifying potential structural and logical discrepancies between the Prolog code and the tax-inconsistency modeling presented in the GitHub repository (Lawsky, n.d). That modeling implemented in Catala and Python with Z3 theorem approver, served as a comparative benchmark for aligning the Prolog model with formally verified representations of §121. This comparative analysis surfaced four major deficiencies in the earlier implementation, each of which was corrected through iterative code refinement. These errors and their corresponding corrections are described in detail in the *Findings* section.

RQ2 Evaluation

After the Prolog model began producing more accurate and replicable results, a set of *validation runners (*small, automated test programs), were developed to evaluate performance. The validation framework comprised multiple runners designed collectively to test four key properties of the model: whether outcomes were correct, internally consistent, deterministic, and capable of autonomously detecting inconsistencies. These evaluation metrics, adapted from prior research in computational legal reasoning and hybrid neuro-symbolic systems (Bench-Capon & Sartor, 2003; Garcez et al., 2019), provided a systematic way to assess whether LLM-assisted formalization could generate legally faithful, logically coherent, and reproducible results.

Finally, to verify the Prolog model's accuracy, a replication step was conducted using Lawsky's Z3 theorem-prover implementation. This framework systematically tests combinations of the statute's undefined parameters, including the temporal unit of the two-year test and the method for calculating the joint numerator, which represents the combined qualifying period used to compute a couple's joint exclusion under §121(c)(2)(A). Because the statute does not specify how to merge each spouse's ownership, use, and prior-exclusion periods, the implementation evaluates alternative rules based on the *minimum*, *maximum*, or *average* of the two spouses' periods. The Prolog model was extended with the same parameterization to replicate these logic-level comparisons and verify whether both systems produce identical patterns of consistency and divergence.

The complete Prolog code, including all validation runners, is provided in *Appendix 3*. The results of all validation experiments are presented in the *Findings* section.

## III. Findings

Research Question One

*Can GPT-4o, when prompted with natural language and Prolog-augmented prompts, reliably detect a known inconsistency in IRC §121?*

Following the procedure outlined in the Research Methodology section above, two types of prompting experiments with GPT-4o were undertaken. Comparative results from these experiments are summarized in *Table 1*.

*Table 1. Comparison of the Results from Two Prompting Approaches.*

| **Prompt without Prolog** | **Output** | **Prompt with Prolog** | **Output** |
|---|---|---|---|
| You are acting as a tax expert. Inconsistency is defined as incompatible demands or permits incompatible actions. You want to isolate any inconsistencies in the attached document (Section 121.pdf) | **P1 - without Prolog**<br><br>Accuracy: 0<br>Rule coverage: 66% | You are acting as a tax expert. You want to isolate any inconsistencies in the attached document (Section_121.pdf) using the attached Prolog rules (no_Facts_Final_Prolog_code_121.pl). Inconsistency is defined as incompatible demands or permits incompatible actions. Provide step-by-step reasoning detailing how you reach any conclusions on inconsistencies in §121 attached. | **P1 – with Prolog**<br><br>Two relevant cases are analyzed.<br>#1. Accuracy Score: 0, Rule Coverage: 66%.<br>#2. Accuracy Score: 0, Rule Coverage: 66%. |
| §121 appears to include two rules that both apply in a particular situation and sometimes give different outcomes. | **P2 - without Prolog**<br><br>Accuracy: 0<br>Rule coverage: 33% | Continue analyzing §121 as a tax expert using the Prolog rules previously attached (no_Facts_Final_Prolog_code_121.pl). §121 appears to include two rules that both apply in a particular situation and sometimes give different outcomes. | **P2 – with Prolog**<br><br>Accuracy Score: 0, Rule Coverage: 66%. |
| §121c(1) seems to have an inconsistency, "*whichever is* | **P3 - without Prolog**<br><br>Accuracy: 1 | §121(c)(1) seems to have an inconsistency, "*whichever is applicable.*" Provide your step-by-step reasoning using the previously attached | **P3 – with Prolog**<br><br>Accuracy: 1 |

| *applicable*." Provide your step-by-step reasoning and conclusion as to whether or not it is and why. | Rule coverage: 100% | Prolog rules (no_Facts_Final_Prolog_code_121.pl). Conclusion as to whether or not it is and how. | Rule coverage: 66% |
|---|---|---|---|

The result from each experiment is summarized below.

*Natural Language Prompting Response Evaluation*

Prompt #1-General Inconsistency Detection- GPT-4o failed to identify the target inconsistency when prompted. The model conducted a systematic analysis covering multiple IRC §121 sections but missed the critical interaction between §121(b)(2)(A), §121(b)(2)(B), and §121(c). While comprehensive, the analysis failed to identify the "whichever applicable" ambiguity. Evaluation = Accuracy Score: 0, Rule Coverage: 66%.

Prompt #2-Guided Rule Conflict Prompt- GPT-4o incorrectly identified a statutory interaction when prompted. The model identified a conflict between §121(b)(3) and §121(c) regarding sales within two years of a prior sale, but this represents a different type of interaction than the target inconsistency. Evaluation = Accuracy Score: 0, Rule Coverage: 33% (wrong sections identified).

Prompt #3 - Specific Whichever is Applicable Prompt- GPT-4o successfully detected the target inconsistency when prompted. The model correctly identified the ambiguity in §121(c)(1) and provided a concrete example demonstrating how partial compliance creates interpretive conflicts between joint and individual exclusion calculations. Evaluation= Accuracy Score: 1, Rule Coverage: 100% (all relevant sections cited and applied).

In sum, GPT-4o presented a limited ability to detect the target inconsistency when prompted with natural language prompting alone. Across three different prompting strategies, the model achieved only a 33% success rate (the percentage of the responses with the accuracy score of 1, with significant variation in the rule coverage score ranging from 33% to 100%). The model's best performance occurred when it was explicitly directed to the specific inconsistency phrase "*whichever is applicable*," rather than through multiple randomized trials of the same prompt.

*Prolog Augmentation Prompt Response Evaluation.*

Prompt #1-with Prolog - General Inconsistency Detection: GPT-4o suggested four different cases of inconsistency. Only two cases not related to the ground truth scenario are analyzed below.

- Case 1 – incorrectly identified a different inconsistency when prompted based on the Prolog code: *joint_return exclusion(Husband, Wife, Property, SaleDate, 500,000) combined_exclusion(...) % fallback to prorated values if joint_return_exclusion fails.* Evaluation= Accuracy Score: 0, Rule Coverage: 66%.

- Case 2 – GPT-4o presented a theoretical statutory conflict between §121(c) and §121(b)(2)(B); specifically, the ambiguity in how partial exclusions due to hardship interact with the fallback formula for joint filers. It recognized that while both rules may apply, the lack of clear integration creates a conflict that satisfies the definition of inconsistency provided in the prompt: "two rules that both apply in a particular situation and sometimes give different outcomes." Evaluation= Accuracy Score: 0, Rule Coverage: 66%.

Prompt #2-with Prolog-Guided Rule Conflict Prompt suggested two different case scenarios of inconsistency. Only a relevant response was analyzed with the evaluation rubric. In this case, GPT-4o failed to detect the inconsistency case described in the ground truth scenario. It applied §121(b)(2)(A) and §121(b)(2)(B) but missed §121(c). Evaluation= Accuracy Score: 0, Rule Coverage: 66%.

Prompt #3 – with Prolog - Specific Whichever is Applicable Prompt: GPT-4o was able to identify the inconsistency problem by concluding: "§121(c)(1) contains a statutory inconsistency arising from the phrase '*whichever is applicable*.'" It creates ambiguity when a taxpayer partially fails the joint exclusion requirements but qualifies for prorated relief under §121(c). The law does not clearly define whether the $250,000 or $500,000 cap should be used in those cases." The model successfully identified and applied §121(c)(1) and §121(b)(2)(A), but it failed to mention §121(b)(2)(B), the fallback provision for individual exclusions. This is a critical omission, as the conflict arises precisely from the ambiguity between applying the joint exclusion cap under §121(b)(2)(A) or the summed individual caps under §121(b)(2)(B) in a partial qualification scenario governed by §121(c). Evaluation= Accuracy Score: 1, Rule Coverage: 66%.

The model's failure to cite §121(b)(2)(B) suggests that it did not fully grasp the statutory mechanics of the conflict. This failure is emblematic of a core weakness in probabilistic models for legal reasoning. The LLM, operating on statistical patterns, identified a surface-level conflict based on the *"whichever is applicable"* language, which it has likely encountered in many training contexts. However, it failed to perform the deeper, structural reasoning required to trace the logical implications of both potential interpretive paths laid out in §121(b)(2)(A) and §121(b)(2)(B). The model identified the symptom but could not fully diagnose the underlying structural problem. This underscores the fundamental unreliability of using LLMs as standalone legal reasoners for high-stakes analysis, as they may generate plausible-sounding but incomplete or systematically flawed justifications.

Research Question Two

*Can a hybrid approach, in which an LLM assists in generating and refining formal Prolog representations of statutory text, produce a deterministic inconsistency detection system that reliably identifies conflicts across diverse use cases?*

As described in the research methodology above, GPT-5 was prompted to identify structural and logical discrepancies between the Prolog code and Lawsky's benchmark frameworks. The comparative analysis revealed the following four major deficiencies:

1. **Missing dual interpretation structure (§121(b)(2)(A) vs. §121(b)(2)(B))**

The Prolog code encoded only one interpretation, the "sum of prorated individual limits, "and could not therefore evaluate the divergence between §121(b)(2)(A) (*joint cap*) and §121(b)(2)(B) (*sum of limitations*). Without both interpretations represented, the model could not detect when the two provisions produced inconsistent results for the same fact pattern. The revised Prolog version introduced a dual interpretation framework that modeled both readings in parallel, enabling the solver to compare their outputs and flag divergence whenever outcomes differed.

2. **Omitted "shortest of three" rule (§121(c)(1)(B))**

Under §121(c)(1)(B), the allowable exclusion must be prorated according to the shortest of three periods: ownership, use, and time since the last exclusion. The baseline code failed to enforce this requirement, applying proration only to the use period. As a result, it overstated eligibility for taxpayers with shorter ownership or prior exclusion intervals. The revised Prolog code corrected this by computing the minimum of all three periods (the "shortest of three" rule) before applying the 24-month test, ensuring the exclusion ratio aligned precisely with §121(c)'s temporal logic.

3. **Misapplied trigger for proration (§121(c))**

The baseline implementation treated the presence of any qualifying reason (such as employment change, health, or unforeseen circumstances) as an automatic trigger for proration, even when the taxpayer already satisfied the full two-out-of-five-year test. This produced incorrect reductions in exclusion amounts, contradicting §121(c)(1)(A), which limits proration to cases where the time requirement has not been met. The revised code explicitly conditioned proration on failure to satisfy the two-year test, ensuring that the reduced exclusion applied only under §121(c) rather than §121(a).

4. **Collapsed joint return conditions (§121(b)(2)(A))**

The baseline version allowed one spouse's ownership and use history to satisfy both spouses' eligibility under §121(b)(2)(A), collapsing distinct statutory conditions (ownership, use, and prior exclusion limits) that apply separately to each taxpayer filing jointly. This masked ambiguity in how the *joint cap* interacts with individual time tests obscuring potential conflicts between §121(b)(2)(A)'s aggregate $500,000 limit and §121(c)(2)(A)'s proportional reduction. The revised implementation introduced explicit joint numerator modes (such as *min of six*, *min three joint*, and *held b2A months*), allowing systematic comparison of alternative joint return interpretations.

**Validation results**

After the errors described above were corrected, GPT-5 was utilized to design and refine a series of *validation runners (as outlined* in the Research Design section). In logic-based systems such as Prolog, a runner is an executable unit that loads specific facts and initiates the reasoning process under controlled conditions.

Two types of runners were implemented to establish the model's internal validity: fixed and mixed.

The fixed runners confirmed that the Prolog model accurately reproduced established benchmark outcomes under controlled statutory scenarios, validating baseline correctness and internal consistency. The mixed runner extended this validation framework by integrating fixed and automated procedures within a single execution sequence. Beginning with correctness and determinism checks, it then embedded the CLP(FD) (Codognet & Díaz, 1996) search method to autonomously generate and evaluate new fact patterns, enabling automated inconsistency discovery across previously untested factual spaces. Together, these runners confirmed that the model satisfies all four core properties of a sound formal reasoning system: correctness**,** internal consistency**,** determinism**,** and autonomous inconsistency detection (Bench-Capon & Sartor, 2003; Garcez et al., 2019).

The results are summarized in *Table 2*, which presents representative input configurations, outputs, and interpretive findings.

*Fixed Runners*

The three fixed-runner tests confirmed that the Prolog implementation reproduced benchmark results with precision and interpretability.
- The Full Qualification Case verified baseline correctness: Both taxpayers met the two-year ownership and use requirements (30 months and 24 months, respectively). As expected, both the *sum of limitations* (§121(b)(2)(B)) and *joint cap* (§121(b)(2)(A)) interpretations produced identical $500,000 exclusions. This verified that the program correctly implemented the full-qualification rule and did not trigger proration when all statutory conditions were satisfied.
- The Ground-Truth Inconsistency Case was successfully reproduced in Prolog, confirming that the model correctly distinguishes between the *sum of limitations* and *joint cap* interpretations under identical facts. This baseline validation established the model's reliability before extending the analysis to automatically generated fact patterns.
- Comprehensive Validation: This multi-stage validation runner confirmed the Prolog model's correctness, consistency, and deterministic behavior across both fixed and automatically generated cases. The Full, Partial (18/24), and Asymmetric (12/24) Qualification Cases reproduced expected outcomes: identical exclusions under full qualification, and persistent divergence when eligibility was incomplete or asymmetric. Repeated runs produced identical results, verifying determinism. The integrated CLP(FD) tests extended the validation to solver-generated fact patterns, which autonomously identified additional divergence cases consistent with the manually tested scenarios. Together, these results established the model's robustness and confirmed that the observed inconsistencies originate from the statute's logic rather than from implementation artifacts.

*Table 2. Summary of Validation Runner Results*

| Runner Type | Name/Case | Input Facts (Months) | Output | Interpretation / Key Findings |
|---|---|---|---|---|
| Fixed | *run_case_no_ inconsistency* | H: 30/30, W: 24/24 | Sum-of-Limits = $500,000; Reduced-$500k | Verified the Prolog model's baseline correctness; both interpretations |
| | (Full Qualification Case) | | = $500,000 → no divergence | produce identical exclusions, confirming correct implementation of the full-qualification rule. |
| Fixed | *run_case_with_inconsistency* (Ground-truth Case Replication) | H: 30/30, W: 12/12 | Sum-of-Limits = $375,000; Reduced-$500k = $250,000 → inconsistency detected | Replicated the ground-truth case by identifying an inconsistency resulting from the phrase "*whichever is applicable*." |
| Mixed | *run_all_validation_tests* (Comprehensive validation) | Cases: 1. H: 30/30, W: 24/24 (Full Qualification) 2. H: 18/18, W: 24/24 (Partial Qualification) 3. H: 30/30, W: 12/24 (Asymmetric Qualification) 4. Repeated (Determinism Check) 5–6. CLP(FD) automated searches | 1. $500,000 = $500,000 → no divergence 2. $437,500 ≠ $375,000 → divergence 3. $375,000 ≠ $250,000 → divergence 5–6. CLP(FD): $489,583 vs. $479,167 → divergence | Executes a six-part validation sequence confirming correctness, internal consistency, determinism, and automated inconsistency detection. |
| Automatic (part of mixed runner) | *print_fact_pattern* (Automated Conflict Discovery) | Dynamic generation through CLP(FD). H: 120/120/120, W: 120/120/23 | Sum-of-Limits = $489,583; Reduced-$500k = $479,167 → inconsistency detected | Autonomously discovers divergence patterns, validating automated inconsistency. detection with the Prolog model. |
| Automatic (part of mixed runner) | *run_joint_prior_table* (Range Validation) | Varies one spouse's prior-exclusion period (1–23 months) while keeping other conditions fixed to test model stability and identify divergence boundaries. | See *Table 3* | Evaluates Prolog model behavior across varying prior-exclusion periods; results differ below 24 months, align at 24 months, and remain stable thereafter, revealing a narrow |

| | | | | "inconsistency zone." |
|---|---|---|---|---|

*Automated Components of the Mixed Runner*

The automated parts of the mixed runner extended the analysis beyond predefined cases, allowing the Prolog model to detect inconsistencies autonomously across a broader set of fact combinations.

Automated Conflict Discovery (*print_fact_pattern*):

This phase marked the transition from manual examples to automatically generated fact patterns within the mixed validation runner. Using CLP(FD), it treated ownership, use, and prior-exclusion periods as variables restricted by statutory conditions. The model autonomously identified a configuration where one spouse's shorter prior-exclusion period reduced the total exclusion below the sum of individual limits. This validated the model's capacity for automatic inconsistency detection, confirming that the conflicts reflect genuine ambiguities in statutory logic rather than outcomes of the manually defined test cases used in the fixed-runner phase.

Automated Range Validation (*run_joint_prior_table*):

This phase extended the automated testing process by systematically varying one spouse's prior-exclusion period from 1 to 36 months while keeping all other eligibility conditions constant. For each configuration, Prolog computed four exclusion totals: the aggregate of individual limits (*SumA*) and three joint-cap interpretations (*min_six*, *min_three_joint*, and *held_b2A_months*).

*SumA* represents the total exclusion under the *sum of limitations* interpretation (§121(b)(2)(B)), calculated as the sum of each spouse's prorated or full $250,000 allowance. *min_six* applies the strictest "minimum-of-six-tests" rule, which evaluates the smallest ownership, use, and prior-exclusion periods across both spouses and their corresponding time-test combinations to determine eligibility. *min_three_joint* uses a hybrid numerator combining the maximum ownership period with the minimum use and prior-exclusion periods, while *held_b2A_months* implements an ownership-gated mode that first checks for joint ownership and then applies the shorter of the use or prior-exclusion months.

The results revealed a distinct shift: outcomes differed below 24 months, aligned at 24 months, and stayed identical thereafter (*Table 3*). This revealed an inconsistency zone concentrated in shorter prior-exclusion intervals, demonstrating that the Prolog model can trace the statute's boundary conditions and reproduce the transition from inconsistency to stability across time.

*Table 3* summarizes how exclusion amounts change as the prior-exclusion period (*P*) varies from 1 to 36 months. The results reveal three distinct behavioral ranges.

For P = 1–23 months, the *SumA* column consistently exceeds all three joint-cap interpretations (*min_six*, *min_three_joint*, and *held_b2A_months*). This means that when one spouse's prior exclusion occurred less than two years earlier, the couple's combined individual entitlements surpass the $500,000 *joint cap*. This divergence creates an inconsistency zone, where §121(b)(2)(A) and §121(c)(2)(A) yield different outcomes.

*Table 3. Prolog-Generated Exclusion Totals Across Varying Prior-Exclusion Periods*

| P | SumA | min_six | min_three_joint | held_b2A_months |
|---|------|---------|-----------------|-----------------|
| 1 | $260,417 | $20,833 | $20,833 | $20,833 |
| 2 | $270,833 | $41,667 | $41,667 | $41,667 |
| 3 | $281,250 | $62,500 | $62,500 | $62,500 |
| 4 | $291,667 | $83,333 | $83,333 | $83,333 |
| 5 | $302,083 | $104,167 | $104,167 | $104,167 |
| 6 | $312,500 | $125,000 | $125,000 | $125,000 |
| 7 | $322,917 | $145,833 | $145,833 | $145,833 |
| 8 | $333,333 | $166,667 | $166,667 | $166,667 |
| 9 | $343,750 | $187,500 | $187,500 | $187,500 |
| 10 | $354,167 | $208,333 | $208,333 | $208,333 |
| 11 | $364,583 | $229,167 | $229,167 | $229,167 |
| **12** | **$375,000** | **$250,000** | **$250,000** | **$250,000** |
| 13 | $385,417 | $270,833 | $270,833 | $270,833 |
| 14 | $395,833 | $291,667 | $291,667 | $291,667 |
| 15 | $406,250 | $312,500 | $312,500 | $312,500 |
| 16 | $416,667 | $333,333 | $333,333 | $333,333 |
| 17 | $427,083 | $354,167 | $354,167 | $354,167 |
| 18 | $437,500 | $375,000 | $375,000 | $375,000 |
| 19 | $447,917 | $395,833 | $395,833 | $395,833 |
| 20 | $458,333 | $416,667 | $416,667 | $416,667 |
| 21 | $468,750 | $437,500 | $437,500 | $437,500 |
| 22 | $479,167 | $458,333 | $458,333 | $458,333 |
| 23 | $489,583 | $479,167 | $479,167 | $479,167 |
| **24** | **$500,000** | **$500,000** | **$500,000** | **$500,000** |
| **25-36** | **$500,000** | **$500,000** | **$500,000** | **$500,000** |

At P = 12 reproduces the ground-truth inconsistency case identified in the fixed-runner tests, serving as the midpoint example within this zone and confirming that the automatically generated results align with manually validated scenarios.

At P = 24 months**,** all columns converge at $500,000, indicating that both spouses now meet the two-year requirement and the statutory interpretations produce identical results. This alignment marks the boundary where the statute transitions from inconsistency to consistency.

For P values beyond 24 months (up to 36 months), all values remain fixed at $500,000 across columns, demonstrating that once the two-year condition is met, the law behaves consistently, and no further divergence occurs. This confirmed that the Prolog system's reasoning aligned not only with human-coded benchmarks but also with independently rediscovered statutory boundaries.

The validation confirmed that the final Prolog implementation met all four target properties of a robust formal reasoning framework. Fixed runners verified correctness and internal consistency under controlled statutory scenarios, while mixed runners demonstrated determinism and, through the integration of CLP(FD), autonomously identified inconsistency patterns consistent with theoretical expectations. Collectively, these results show that the model is legally faithful, computationally stable, and capable of reproducing benchmark outcomes while systematically exploring new fact spaces. Building on these results, the next phase cross-verified the Prolog outputs against Z3 to test whether both frameworks revealed identical divergence patterns, thereby establishing external validity.

*Cross-Validation with Z3 Theorem-Prover Results*

This validation step externally verified the Prolog model against the Z3 theorem-prover implementation. Lawsky's code systematically varies two parameters that §121 leaves unspecified: the temporal unit for the two-year test (days, months, or years) and the joint-numerator rule for combining spousal periods (minimum, maximum, or average). The output file (*checkallunits.txt*) reports, for each parameter combination, whether the *sum of limitations* and *joint cap* interpretations under identical facts produces equal or divergent exclusion amounts.

The Prolog model was extended to mirror these settings. When tested under identical configurations, it reproduced the same convergence and divergence pattern across all units of time and spousal-combination modes:

- Divergence appeared under the *minimum* and *maximum* joint-numerator rules (e.g., days/max = $251,027 vs $251,370; years/max = $375,000 vs $500,000).
- Convergence occurred consistently under the *average* rule, where both interpretations produced identical outcomes (e.g., years/average = $375,000).

This exact replication of Z3's logic-level outcomes confirms that both systems independently identified the same instability boundary within §121. It demonstrates that the Prolog implementation, guided by GPT-5, not only achieved internal determinism but also external consistency with a formally verified theorem-prover model.

This research demonstrates that a hybrid approach combining LLM-assisted formalization with symbolic logic can produce a deterministic inconsistency detection system that reliably identifies statutory conflicts in tax law across diverse use cases.

By encoding §121's competing provisions: the *sum of limitations* and *joint cap* interpretations into executable predicates, the Prolog model exposes the statute's internal ambiguity through explicit, reproducible reasoning steps. Using GPT-5 to assist in code refinement, the Prolog implementation evolved from basic eligibility tests into a fully parameterized reasoning system capable of handling diverse ownership, use, and prior-exclusion combinations.

The validation runners confirmed the model's internal reliability across multiple scenarios, while cross-replication with the Z3 theorem prover established its external soundness. The replicated outputs matched the divergence patterns documented in Lawsky's *checkallunits.txt* file, confirming that both systems independently identified the same logical boundary of inconsistency within §121.

The Prolog model revealed an inconsistency zone within §121, offering computational insight into why the IRS adopted the sum-of-limitations approach in Publication 523. This finding suggests that the IRS's position, though not explicitly codified, aligns with the interpretation that preserves stability across the statutory parameter space.

This study represents a conceptual advance in hybrid neuro-symbolic reasoning for legal interpretation. With proper human guidance, LLMs can translate statutory logic into structured code capable of detecting internal inconsistencies, bridging the gap between natural-language reasoning and formal verification. The Prolog implementation illustrates how LLM-assisted logic programming can surface statutory ambiguities, evaluate competing interpretations, and support deterministic inconsistency detection in legal texts.

## IV. Discussion

Our findings establish a clear hierarchy for tax law consistency detection. A deterministic symbolic reasoner should anchor the analysis, with LLMs providing valuable assistance in prompt augmentation and rule formalization. This represents a significant departure from approaches that position LLMs as primary reasoning engines for legal analysis. Instead, LLMs function most effectively as assistive translators and augmenters, enabling rapid conversion of legal text into structured logic that symbolic systems can execute, audit, and verify.

Empirical Evidence for Symbolic Superiority

The performance gap between approaches is stark and consistent:
- Symbolic reasoning (Prolog) deterministically produced the correct outcome in all tested scenarios, with fully auditable logic (100% success rate).
- Probabilistic reasoning (GPT-4o) was prompt-sensitive: only one of three prompts (33% success rate) elicited the correct identification of inconsistency, and reasoning quality varied considerably.

The Value and Limitations of LLMs

Although LLMs failed as a primary reasoning engine, their contributions to the formalization of legal rules were substantial:

LLM Contributions:
- Prompt enhancement.
- Rapid translation of statutory language into formal logic structures produced usable Prolog code in SWISH, though with many errors and strong dependence on prompting. Nevertheless, it provided a foundation for iterative refinement.
- Identification of relevant statutory sections and their relationships.
- Generation of test scenarios and edge cases.
- Debugging assistance for symbolic rule refinement.

LLM Limitations:
- Inconsistent performance across similar prompts.

- Inability to systematically apply complex rule interactions.
- Lack of genuine reasoning.
- Lack of audit trails for reasoning steps.

Implications for Legal Technology Development

These findings suggest a clear pipeline for legal AI systems:

1. LLM-Assisted Formalization

- LLMs can serve as a viable starting point to translate natural language statutes into formal representations.
- LLMs can assist in rapid prototyping and rule generation.
- LLMs provide assistance for debugging and refinement of symbolic rules and program logic.

2. Symbolic Execution and Verification

- Execute formal rules in deterministic symbolic environments.
- Generate complete audit trails for legal reasoning.
- Ensure reproducible, explainable results.

Practical Applications

This advanced approach has direct applications for:

- Legislative drafting: Automated detection of conflicts before statute enactment.
- Tax compliance: Systematic checking of complex regulatory interactions.
- Legal education: Transparent demonstration of statutory reasoning processes.
- Regulatory analysis: Comprehensive consistency checking across legal frameworks.

**Limitations**

Our research has several important limitations:

1. Testing was focused on a single IRC section and one type of inconsistency.
2. Only GPT-4o and GPT-5 were tested in distinct stages of this study; other LLMs might perform differently.
3. The evaluation scale has a limited number of test scenarios and a single human evaluator, which may have introduced bias.
4. Prompts were iteratively refined with guidance from a legal expert. While this improved specificity, it may have introduced bias into the prompt design.
5. Testing was conducted via the ChatGPT web interface to align with typical practitioner use by avoiding the requirement for additional technical setup. The trade-off is limited control over system-level configurations, such as temperature settings, token constraints, and the structure of internal prompts.

6. Experiments for RQ2 were conducted sequentially with two different models, GPT-4o and GPT-5, reflecting the rapid development of LLMs during the research period. The refinement stage, therefore, represents a methodological evolution rather than a strict replication. Results might differ if the entire experiment were re-run exclusively with GPT-5.

**<u>Future Research Directions</u>**

This study opens several promising avenues for future investigation:

- Scaling the methodology. Future research can test whether this hybrid approach can formalize and analyze entire subchapters of the IRC and/or other complex regulatory frameworks. Such studies would validate the scalability and generalizability of our pipeline beyond single-section analysis.

- Expand testing to additional IRC sections. The current research focused on a single, known inconsistency in IRC §121. Future research can apply the same hybrid methodology to other sections of the Internal Revenue Code to check the robustness of this solution. This would help determine if the neuro-symbolic approach is effective for a broader range of legal inconsistencies, including several types of conflicts identified by Donelson (2019), such as reconcilable inconsistencies or conflicts between justification and execution. A human-in-the-loop approach is essential for guiding LLMs, using ground-truth cases to highlight inconsistency issues and to support the formalization of legal rules.

- Discovering unknown inconsistencies. Another application involves moving from verifying known inconsistencies to discovering unknown ones. This would require developing systematic Prolog queries that can test for logical contradictions such as conflicting obligations, circular references, or impossible conditions across large corpora of formalized rules. Such a capability would transform this from a verification tool into a discovery engine.

- Integration with legislative drafting. The ultimate application lies in integrating this methodology into the legislative drafting process itself. A system providing real-time feedback to legislative counsel, flagging potential inconsistencies during the writing process, could prevent these issues from becoming law. Research into the human-computer interaction aspects of such integration would be essential for practical deployment.

- Test knowledge graph frameworks to enable an LLM to reason over statutory hierarchical relationships. Future work should explore the use of knowledge graphs to improve LLMs reasoning abilities. Legal documents, particularly the tax code, are inherently hierarchical, with rules and exceptions nested within subsections and paragraphs. A knowledge graph represents statutory provisions as nodes representing sections and edges, which denote relationships like "is a subsection of, "modifies," or "references." This would allow an LLM to perform more accurate and structured reasoning by navigating the graph, thereby overcoming its challenges with deep, hierarchical processing over long text.

- Compare performance across different LLM models and architectures. The study's LLM testing was limited to GPT-4o and GPT-5. As new LLMs and architectures, such as DeepSeek R1 and OpenAI o3-mini, are developed with improved reasoning capabilities, it is crucial to compare their performance in detecting legal inconsistencies. This comparative analysis would help identify which models are best suited for legal formalization and prompt augmentation tasks within the hybrid framework.

**<u>Conclusions</u>**

The potential benefits of deploying large language models (LLMs) in tax law applications are substantial. Improving compliance, the avoidance of potentially unfair outcomes, and as an aid in drafting consistent legislation are important benefits. However, empirical research on tax-specific LLM applications remains sparse.

This research demonstrates that while LLMs provide valuable assistance in translating statutory language into formal rules, deterministic symbolic reasoning systems are essential for reliable statutory inconsistency detection. This finding suggests a complementary division between probabilistic LLMs (for formalization and hypothesis generation) and deterministic symbolic reasoners (for verification and execution).

- Detecting statutory inconsistencies is a legal reasoning problem that involves both natural-language processing and formal logic. The finding that GPT-4o succeeded in only one of three prompting strategies (a 33% prompt success rate) suggests that it struggles to reliably detect inconsistencies on its own. Its reasoning becomes markedly more reliable when augmented with structured chain-of-thought prompting.
- These findings suggest that CoT prompting improves GPT-4o's response accuracy when directed toward a known inconsistency in the tax law, but the model struggles to detect such an inconsistency independent of human-in-the-loop expert-provided guidance. Prompts that included the phrase "*whichever is applicable*," led to correct responses in both baseline and Prolog-augmented scenarios. However, Prolog rule augmentation did not significantly enhance GPT-4o's reasoning performance when embedded within prompts, even though the Prolog rules affected the reasoning process of GPT-4o. In contrast, Prolog rules tested independently in the SWISH environment demonstrated a greater capacity to detect logical inconsistencies, confirming the importance of symbolic reasoning tools in legal analysis.
- This research also demonstrates the potential of using LLMs to formalize statutory language into Prolog rules for computational analysis. This opens a path for AI-assisted legal formalization that can support tasks such as inconsistency detection and compliance verification. It highlights how LLM-generated Prolog rules can be refined and tested in symbolic logic environments such as SWISH**.** This process illustrates the effectiveness of combining symbolic Prolog logic with generative AI tools such as GPT-4o and GPT-5 in legal reasoning research. The approach demonstrates the potential to scale such testing across a broader range of factual situations, offering a systematic and explainable method for identifying statutory inconsistencies and supporting legal compliance analysis. This process can be extended to generate synthetic data for further model refinement specific to various legal tasks.

- Finally, this research contributes to the development of a structured evaluation framework for assessing both the accuracy and reasoning quality of LLM-generated legal outputs. The rubric developed can be further refined to evaluate legal responses in a more automated and scalable manner.

**Funding:** This work was supported, in part, by the National Science Foundation under Award Number 2423235.

**Data Availability Statement:** Data and code supporting this research are available from the corresponding author on request.

**Acknowledgments:** We thank our reviewers for their time and invaluable input: Sarah B. Lawsky, L.B. Lall and Sumitra Devi Lall Professor of Law and Co-Director, Innovation Law and Technology, University of Illinois College of Law, and Jay A. Soled, Distinguished Professor and Chair, Department of Accounting and Information Systems, Rutgers University Business School.

**Conflicts of Interest:** The authors declare no conflict of interest.

**Replicability Statement:** All Prolog code and test cases are available in the GitHub, enabling complete replication of symbolic reasoning experiments. The deterministic nature of Prolog execution ensures identical results across different runs and evaluators.